\documentclass[conference]{IEEEtran}
\IEEEoverridecommandlockouts
\usepackage{cite}
\usepackage{times}  
\usepackage{helvet} 
\usepackage{courier}  
\usepackage[hyphens]{url}  
\usepackage{graphicx} 
\usepackage{graphicx}  
\usepackage{amsmath,graphicx}

\usepackage{amsmath}
\usepackage{xcolor,soul,framed} 
\usepackage{xspace}
\usepackage{array}
\usepackage{eqparbox}
\usepackage{url}
\usepackage{longtable}
\usepackage{lipsum}
\usepackage{blindtext}
\usepackage{makecell}
\usepackage{mathtools}
\usepackage{commath}
\usepackage{multirow}
\usepackage{tabularx}
\usepackage[normalem]{ulem}
\usepackage{xspace}
\usepackage{algorithm}
\usepackage{algpseudocode}
\usepackage{amsfonts}
\usepackage{placeins}
\usepackage{amssymb}
\usepackage{textcomp}
\usepackage[pagebackref=true,breaklinks=true,colorlinks,bookmarks=false]{hyperref}
\hypersetup{
    colorlinks=true,
    filecolor=magenta,      
    urlcolor=magenta,
}

\DeclareMathOperator{\LSTM}{LSTM}
\def\BibTeX{{\rm B\kern-.05em{\sc i\kern-.025em b}\kern-.08em
    T\kern-.1667em\lower.7ex\hbox{E}\kern-.125emX}}
    
\def\BibTeX{{\rm B\kern-.05em{\sc i\kern-.025em b}\kern-.08em
    T\kern-.1667em\lower.7ex\hbox{E}\kern-.125emX}}
    
\usepackage{fancyhdr}
\begin{document}

\bstctlcite{IEEEexample:BSTcontrol}

\title{Deep Recurrent Semi-Supervised EEG Representation Learning for Emotion Recognition}


\author{\IEEEauthorblockN{Guangyi Zhang, Ali Etemad}
\IEEEauthorblockA{\textit{Dept. ECE and Ingenuity Labs Research Institute} \\
\textit{Queen's University, Kingston, Canada}\\
\{guangyi.zhang, ali.etemad\}@queensu.ca}}

\maketitle
\thispagestyle{fancy}

\begin{abstract}
EEG-based emotion recognition often requires sufficient labeled training samples to build an effective computational model. Labeling EEG data, on the other hand, is often expensive and time-consuming. To tackle this problem and reduce the need for output labels in the context of EEG-based emotion recognition, we propose a semi-supervised pipeline to jointly exploit both unlabeled and labeled data for learning EEG representations. 
Our semi-supervised framework consists of both unsupervised and supervised components. The unsupervised part maximizes the consistency between original and reconstructed input data using an autoencoder, while simultaneously the supervised part minimizes the cross-entropy between the input and output labels. We evaluate our framework using both a stacked autoencoder and an attention-based recurrent autoencoder. We test our framework on the large-scale SEED EEG dataset and compare our results with several other popular semi-supervised methods. Our semi-supervised framework with a deep attention-based recurrent autoencoder consistently outperforms the benchmark methods, even when small sub-sets ($3\%$, $5\%$ and $10\%$) of the output labels are available during training, achieving a new state-of-the-art semi-supervised performance. 
\end{abstract}


\section{Introduction}
Human emotion is defined as a mental state related to the Central Nervous System (CNS) \cite{dalgleish2004emotional}. It is caused by conscious or unconscious perception of emotion-related stimuli and often leads to changes in our physical and psychological processes \cite{panksepp2004affective}. Emotions have important influence on rational mechanisms such as decision making, perception, and other cognitive processes \cite{picard2000affective}. It is thus important to develop analytical or data-driven computational models for understanding, communicating, and responding to human emotions \cite{picard2000affective}. 

Many non-invasive physiological signals have been employed for emotion recognition. Examples include muscle electrical activity with Electromyography \cite{jerritta2011physiological}, electrical signals from the heart through Electrocardiography  \cite{sarkar2019classification}, and measuring skin conductance via Galvanic Skin Response \cite{ross2019toward}. Among such technologies, Electroencephalography (EEG) has the potential to demonstrate the highest fidelity due to the direct measurement of signals and information from the brain.

\begin{figure}
    \begin{center}
    \includegraphics[width=0.75\columnwidth]{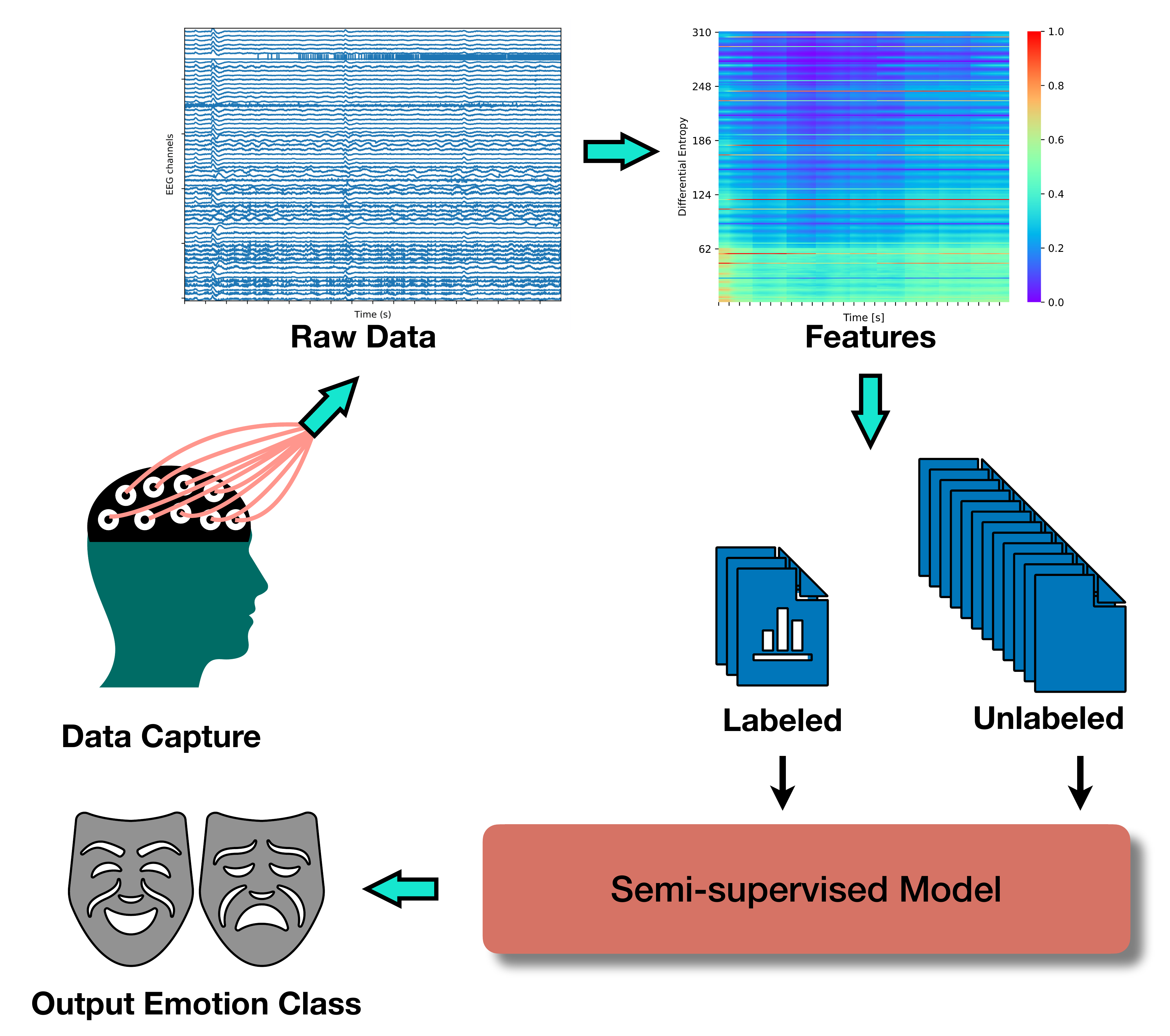} 
    \caption{An overview of the training phase of our semi-supervised EEG-based emotion recognition pipeline is presented.}
    \label{fig: overview}
    \end{center}
\end{figure}

EEG are non-stationary signals, often with high dimensionality. Moreover, EEG often contain various artefact and noise during the recordings, resulting in lower signal-to-noise ratio. Therefore, many supervised pipelines with deep learning techniques have been recently employed to learn the discriminative information from EEG time-series for various tasks such as hand movement classification \cite{zhang2019classification}, motor imagery classification \cite{zhang2020rfnet}, and driver vigilance estimation \cite{zhang2019capsule}, achieving state-of-the-art results. These Fully Supervised Learning (FSL) pipelines highly rely on large amounts of labeled training samples to overcome challenges such as high dimensionality, mixing effect among recorded channels, and existence of noise and artifacts in the data.

FSL method generally require enough labels for effective training. However, annotating and labeling EEG is often very hard and expensive, and requires the help of experts. Especially for the task of emotion recognition, EEG labeling may require both subject self-evaluation as well as objective evaluation (performed by experts) \cite{zheng2015investigating}. Moreover, the performance of FSL methods often degrade when labeled samples are insufficient compared to when all the samples are labeled.

In this paper, to rely only on small amounts of labeled data, we propose a Semi-Supervised Learning (SSL) pipeline to recognize emotions using a deep recurrent autoencoder (AE). This AE is trained in an unsupervised fashion and its encoder component is followed by a simple two-layer classifier which is trained in a supervised manner simultaneously with the AE.
Our approach can operate effectively by learning mainly from features extracted from unlabeled data while requiring only small subsets of the available labels for receiving the necessary supervision. An overview of the training phase of our framework is depicted in Figure \ref{fig: overview}. Training is performed jointly using the labeled and unlabeled data with the total loss incorporating both the supervised and unsupervised losses. We perform extensive experiments and show that our approach is able to provide better performance than other SSL benchmarks when small amounts of labeled data are used for training. In summary, our contributions are as follows. 
(\textbf{1}) 
We propose a semi-supervised approach for EEG-based emotion recognition to leverage large amounts of unlabeled and along with small amounts of labeled data. 
(\textbf{2}) 
We conduct extensive experiments by re-implementing high quality SSL pipelines, which are equipped with deep learning techniques. 
(\textbf{3}) 
We show the robustness of our SSL method by consistently achieving better results than the other methods when very few labeled samples are used ($3\%$, $5\%$ and $10\%$), achieving new state-of-the-art . 


\section{Related Work}
\subsection{Emotion Recognition with EEG}
\textbf{Classical Machine Learning.}
Feature extraction is a key step in EEG-based emotion recognition given the non-stationarity and non-linearity of EEG signals. As a result, extraction of discriminative EEG features from large-scale EEG data is essential prior to training classifiers. Most existing works extract features from five brain waves, notably delta, theta, alpha, beta, and gamma bands \cite{alarcao2017emotions}. The feature extraction methods include short-time Fourier transform, wavelet transform, power spectrum density, statistical, common spatial filter, and others \cite{alarcao2017emotions}. Recently, Differential Entropy (DE) has been used as a powerful feature extraction method suitable for many machine learning algorithms in emotion recognition \cite{zheng2015investigating}.

Machine learning classifiers play a fundamental role in feature-space learning for emotion recognition, and many pipelines have been built to map extracted features to emotion classes \cite{alarcao2017emotions}. Various algorithms have been used in conjunction with DE features. Canonical correlation analysis has been employed to model a linear relationship between DE features and emotional labels \cite{zheng2016multichannel}. Machine learning classifiers including Support Vector Machine (SVM), \textit{k}-Nearest Neighbor, Logistic Regression, and Random Forest have been employed to explore non-linear relationships between DE feature vectors and affective labels \cite{zheng2015investigating,li2019novel}.

\textbf{Deep Learning.}
In the context of EEG learning, deep learning techniques have been used to explore more powerful and task-relevant features than conventional machine learning algorithms. Deep Neural Network (DNN) has been used to improve the learning process using multiple hidden layers \cite{zheng2015investigating}. A pipeline built with a Convolutional Neural Network (CNN) has been used to automatically extract features from raw EEG signals \cite{tripathi2017using}. Capsule network has been employed to learn spatio-temporal representations, achieving state-of-the-art results for the task of vigilance estimation \cite{zhang2019capsule}.

Recurrent Neural Networks (RNN) have been successfully utilized for analysis of natural language sequences \cite{cho2014learning}, image sequences \cite{sepas2020long}, and bio-signals including EEG \cite{zhang2019classification} to take advantage of their ability to learn short-term dependencies in sequential data. However, conventional RNNs face the vanishing/exploding gradient problems when sequences are long \cite{zhang2019classification}. An enhanced type of RNN capable of learning both long- and short-term dependencies, namely Long Short-Term Memory (LSTM) networks, have also been applied to EEG signals \cite{zhang2016continuous}. Other than temporal information processing, deep RNNs have also been used to learn spatial dependencies of EEG electrodes on the scalp \cite{zhang2018spatial}.

Very recently, graph neural networks have been employed to discover topological information of EEG channels and further establish graph connections, achieving very promising results in an emotion recognition \cite{song2018eeg,zhong2020eeg}. To further explore topological structures, a Riemannian-based network was proposed to explore the spatial correlation information of EEG recordings that lie in the Riemannian manifold, demonstrating very strong performance in multiple EEG-based tasks including emotion classification \cite{zhang2020rfnet}.

\subsection{Semi-Supervised Learning}
\textbf{Unsupervised Pre-training.}
Unsupervised pre-training methods equipped with deep learning architectures have been successfully used in computer vision \cite{bulat2021pre}. In \cite{erhan2010does}, it was argued that unsupervised pre-training acts as regularization which helps the model converge towards a local minimum during the main supervised training. Generally, unsupervised pre-training includes a Stacked AutoEncoder (SAE) which is used with available training data to update the model weights until minimum unsupervised training loss is reached. Next, the pre-trained encoder is followed by a classifier which is trained with labeled data in a supervised manner (also called fine-tuning) \cite{van2020survey}. Similarly, a pipeline consisting of a Deep belief Network (DBN) for unsupervised pre-training, followed by logistic regression for supervised fine-tuning has been used for EEG-based affective state recognition \cite{xu2016affective}.

\textbf{Pseudo Labeling.}
An alternative approach to the above is to develop pipelines that use labeled and unlabeled data simultaneously to train the network. 
In \cite{lee2013pseudo}, a model is first trained on labeled data, and then used to generate pseudo-labels for the unlabeled data. The model is then retrained on labeled data and pseudo-labeled data together. This approach outperformed several FSL methods with machine learning and deep learning backbones when small amounts of labeled data were available.

\textbf{$\mathbf{\Pi}$ Model.}
The performance of the pseudo-labeling method mentioned above \cite{lee2013pseudo} has been reported unstable when used on several image datasets \cite{oliver2018realistic} when very few samples are labeled. The $\Pi$ model was proposed to apply consistency regularization on both unlabeled and labeled data, leading the network to be more robust to noisy or perturbed inputs \cite{laine2016temporal,athiwaratkun2018there}. To do so, the $\Pi$ model first applies stochastic augmentation with additive noise on the input as well as dropout on the network. Next, it trains the network to obtain two different outputs for each input sample and minimizes the distance between two outputs. Thus, the network is encouraged to make similar prediction outputs given the same input with different random noise \cite{laine2016temporal}. This approach achieves better results than pseudo-labeling with very limited amount of labeled samples \cite{laine2016temporal,oliver2018realistic}.

\textbf{Temporal Ensembling.}
The $\Pi$ model trains slowly since the network trains twice for each input sample. Also, the network outputs are not stable due to the random noise applied on the inputs and the dropout in the network \cite{laine2016temporal}. To tackle these challenges, temporal ensembling was proposed to train the network only once for each input and aggregate the network output from previous epochs into an ensemble \cite{laine2016temporal}. Consequently the framework trains faster and the network output is more stable \cite{laine2016temporal}.

\begin{figure*}
    \begin{center}
    \includegraphics[width=0.75\textwidth]{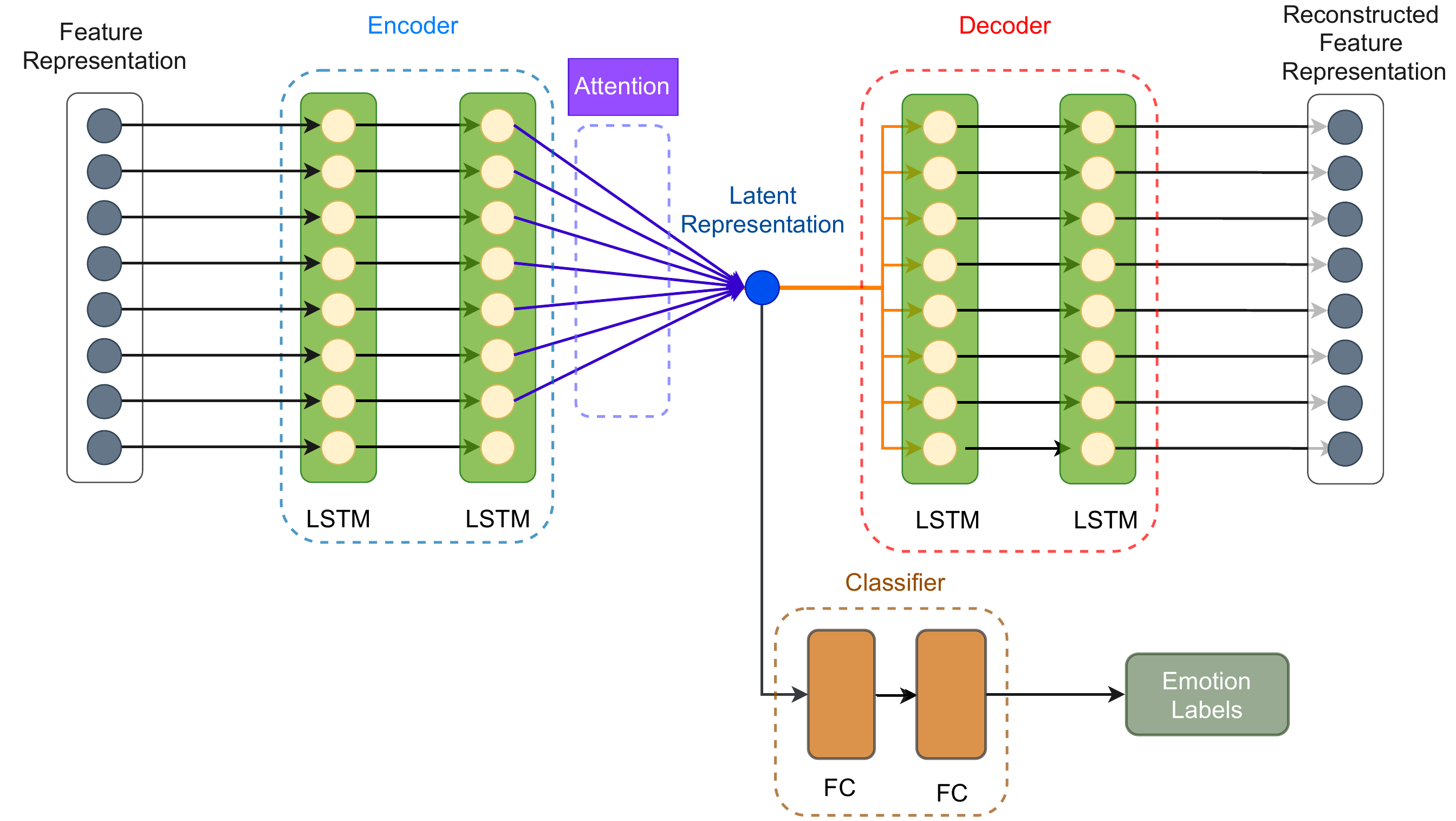} 
    \caption{The proposed semi-supervised model using an attention-based recurrent autoencoder.}
    \label{fig: ALSTM-AE}
    \end{center}
\end{figure*}

\textbf{Mean Teacher.}
Temporal ensembling aggregates the previous model outputs for better model performance. However, it has considerable space requirements to save the previous models' outputs. Moreover, the ensemble outputs update slowly (only once per epoch), resulting in poor performance when the training data size is large \cite{tarvainen2017mean}. 
The mean teacher method was proposed to address these problems through using ensemble model weights instead of ensemble model outputs \cite{tarvainen2017mean}. Specifically, this framework consists of two networks with the same architecture, namely student and teacher. The student model plays the same role as in temporal ensembling, which is trained only once for each input \cite{tarvainen2017mean}. The teacher model is not trained, but instead uses the ensemble model weights from the student which updates them frequently (in each training step). 
Consequently, instead of minimizing the difference between the network output and the ensemble network output as in temporal ensembling, the mean teacher method minimizes the distance between the outputs of the student and the teacher networks.

\section{Our Approach}
We aim to build a semi-supervised pipeline capable of using only small amounts of labeled data while mainly relying on unlabeled data for EEG representation learning. In the following sections, we first give a brief overview of feature space. Next, we introduce our proposed solution based on an Attention-based Recurrent AutoEncoder (Att. RAE).

\textbf{Feature Space.}
We perform the following pre-processing steps according to \cite{zhang2020rfnet}. First, EEG time-series were downsampled from $1$ \textit{KHz} to $200$ \textit{Hz}. Next, a bandpass filter was applied to minimize noise and artifacts outside the frequency range of $[0.5-70]$ \textit{Hz}. Then, a notch filter was applied to reduce the power line effect around $50$ \textit{Hz}. Signal normalization was followed to ensure that the amplitude of recordings from different subjects fall into the same range of $[-1, 1]$. 

Following pre-processing, we extract DE features from each consecutive $1$-second windows with no overlap on each $8$-second EEG segment. We assume the signals follow a Gaussian distribution, and therefore compute the DE feature using Eq. \ref{equation: de} \cite{zheng2015investigating}. 
\begin{equation}\label{equation: de}
\small
DE = \frac{1}{2} \log{2\pi e \sigma^{2}}, \hspace{3mm} x\sim N(\mu, \sigma^{2}).  
\end{equation}
We extract DE features from five EEG bands for each of the $62$ EEG channels, yielding a total of $310$ features \cite{zheng2015investigating}.

\textbf{Deep Recurrent AutoEncoder.}
We propose a recurrent autoencoder \cite{li2020extraction} consisting of an encoder which maps the input onto a latent space, followed by a decoder to reconstruct the original input using the latent representation. The encoder is made up of two layers of LSTM, followed by a soft attention mechanism \cite{zhang2019classification, zhang2020rfnet}, as illustrated in Figure \ref{fig: ALSTM-AE}. The first LSTM layer employs $8$ cells to process EEG features extracted from each window sequentially. Specifically, DE features (denoted as ${\{x_t\}}_{t=1}^8$) from $t^{th}$ window are fed into the corresponding $t^{th}$ cell of the first LSTM layer, as shown in Eq. \ref{h_t}. The second LSTM layer inherits and further processes the hidden state outputs (${{h_t}^{enc1}}$) from the previous LSTM layer and computes hidden state outputs in each cell (${h_t}^{enc2}$), as shown in Eq. \ref{h_t'}. 
\begin{equation}\label{h_t}
\small
{h_t}^{enc1} = \LSTM({x_t})
\end{equation}
\begin{equation}\label{h_t'}
\small
{h_t}^{enc2} = \LSTM({h_t}^{enc1})
\end{equation}

To obtain a well-performed output representation from the recurrent network, it is necessary to evaluate the importance of the output of each recurrent step \cite{zhang2019classification}. Therefore, we apply a soft attention mechanism following the last LSTM layer, enabling attention weights $\alpha_t$ to be assigned to each LSTM cell's output ${h_t}^{enc2}$ as shown in Eq. \ref{u_t} and \ref{alpha_t}, where $W$ and $b$ are the trainable weights and biases. Following, $v$ is calculated using Eq. \ref{v_t} as the output of the attention mechanism as the sum of the hidden outputs of all LSTM cells multiplied with the assigned trainable attention scores. More discriminative features can hence be obtained by optimizing the attention weights \cite{zhang2020rfnet}. 
\begin{equation}\label{u_t}
\small
{u_i} = tanh({W} {h_t}^{enc2} + {b})
\end{equation}
\begin{equation}\label{alpha_t}
\small
{\alpha_t} = \frac{exp({u_t})}{\sum_k{exp({u_k})}}
\end{equation}
\begin{equation}\label{v_t}
\small
{v} = \sum_t{{\alpha_t}{h_t}^{enc2}}
\end{equation}

In the decoder, we employ two LSTM layers similar to the encoder. The first LSTM layer feeds the obtained latent representation ${v}$ into each of the $8$ LSTM cells, as shown in Eq. \ref{h_t_decoder}. Then, the last layer further processes the output hidden states ${h_t^{dec1}}$ and computes the output of each cell as shown in Eq. \ref{h_t'_decoder}. 
We denote the reconstructed input from our recurrent autoencoder (Att. RAE) as ${\hat{x}_t}$.
\begin{equation}\label{h_t_decoder}
\small
{h_t}^{dec1} = \LSTM({v})
\end{equation}
\begin{equation}\label{h_t'_decoder}
\small
{\hat{x}_t} = \LSTM({h_t}^{dec1})
\end{equation}

In our LSTM settings, each layer contains $256$ hidden units representing the size (dimension) of each cell's output vector. 

\textbf{Classifier.}
As shown in Figure \ref{fig: ALSTM-AE}, we also use a classifier which is fed the latent representation obtained by the autoencoder (after the encoder component) for the supervised learning part of our method. The classifier consists of two Fully Connected (FC) layers with $64$ and $3$ units, respectively, along with ReLU activations. A dropout rate of $0.5$ is applied on the first FC layer.

\textbf{Semi-Supervised Algorithm.} 
Going forward, we denote the entire training data as $D$, the labeled data as $D_l$, and the unlabeled data as $D_{ul}$. Our loss function consists of two parts, namely unsupervised loss ($\mathcal{L}_u$) and supervised loss  ($\mathcal{L}_s$). In the unsupervised component, we minimize the difference between the original and reconstructed input as obtained by the autoencoder for $D_{ul}$ and $D_{l}$ using MSE loss as in: 
\begin{equation}\label{equation: ul}
\small
\mathcal{L}_{u} = \frac{1}{B_{ul}}\sum_{i=1}^{B_{ul}} ||x_i'-\hat{x}_i'||_2^2 +\frac{1}{B_{l}}\sum_{i=1}^{B_l} ||x_i-\hat{x}_i||_2^2,
\end{equation} 
where $x, {x'}$ are the input features of the labeled and unlabeled data and ${\hat{x}}$. ${\hat{x}}'$ are the corresponding reconstructed input features computed by the autoencoder for labeled and unlabeled data. $B_{ul}$ and $B_l$ are the batch sizes of $D_{ul}$ and $D_l$. 

In the supervised component, we employ cross-entropy loss $\mathcal{L}_s$ for the labeled data only as in: 
\begin{equation}\label{equation: l}
\small
\mathcal{L}_{s} = - \frac{1}{B_l}\sum_{i=1}^{B_l}\log{[f_{\theta_2}'(\mathcal{Z}_{l})]}y_i,
\end{equation} 
where $\theta_1,\theta_2$ are the parameters of the autoencoder and classifier respectively. $\mathcal{Z}_{l}$ is the latent representation learned by the autoencoder with the input of the labeled data $x$. $f_{\theta_2}'$ is the classifier, and $y$ are the labels for $x$.

To perform experiments when few samples are labeled, we randomly choose a very small portion of $D$ as our $D_l \ll D$, and treat the remaining samples as our $D_{ul}$. We then split $D_l$ and $D_{ul}$ equally into $R$ batches, respectively, such that the ratio of $B_l/B_{ul}$ remains the same in each batch. The losses are updated once per batch, as shown in Algorithm \ref{Algorithm}.

During the joint training, we purpose to slowly increase the weight of the unsupervised loss term for accelerating model convergence \cite{laine2016temporal}. To do so, we apply the ramp-up function presented in Eq. \ref{equation: ramp} on the unsupervised loss where coefficient $\eta(t)$ increases as the training epoch $t$ increases \cite{laine2016temporal}. 
\begin{equation}\label{equation: ramp}
\small
\eta(t) = 20\exp{[-5(1-\frac{t}{T})^2]}.
\end{equation}
Accordingly, the total loss is calculated as follows:
\begin{equation}\label{equation: L}
\small
\mathcal{L} = \eta\mathcal{L}_{u} + \mathcal{L}_{s}.
\end{equation}

\begin{algorithm}[!t]
\caption{Semi-Supervised Learning} \label{Algorithm}
\begin{algorithmic}[1]
\Require
    \State $ D_l \lor D_{ul} \subseteq D$ , $D_{l}\land D_{ul} = \emptyset$
    \State $x_i, y_i$: $i^{th}$ data sample and label of $D_l$
    \State ${x'}_i$: $i^{th}$ data sample of $D_{ul}$
    \State $R$: total iterations for mini-batch training
    \State $f_{\theta_1}$, $h_{\theta_1}$, $g_{\theta_1}$ and $f_{\theta_2}'$: encoder, attention, decoder and classifier
    \State $\mathcal{Z}=f_{\theta_1}(x)$: latent representation learned by autoencoder

\Ensure
    \State $D_{l}= D_{l}^1 \lor D_{l}^2 \lor ... \lor D_{l}^R$ \Comment{subset of $D_l$ in batch}
    \State $D_{ul}= D_{ul}^1 \lor D_{ul}^2 \lor ... \lor D_{ul}^R$ \Comment{subset of $D_{ul}$ in batch}
      
\For {t in [1, T] epochs} 
    \State update $\eta(t)$ (Eq. \ref{equation: ramp}) \Comment{ramp-up coefficient}  
\For {r in [1,R] iterations}
    \State $\mathcal{Z}_{ul} \leftarrow h_{\theta_1}(f_{\theta_1}(x_i')$), $x_i' \in D_{ul}^r$ 
    \State $\mathcal{Z}_{l}  \leftarrow h_{\theta_1}(f_{\theta_1}(x_i)$), $x_i \in D_{l}^r$
    \State $\hat{{x'}} \leftarrow g_{\theta_1}(\mathcal{Z}_{ul})$, $\hat{x} \leftarrow g_{\theta_1}(\mathcal{Z}_{l})$
    \State update $\mathcal{L} = \eta\mathcal{L}_{u} + \mathcal{L}_{s}$ (Eq. \ref{equation: L}) \Comment{total loss}
    \State update $\theta_1$, $\theta_2$ 
\EndFor 
\EndFor
\end{algorithmic}
\end{algorithm}

\section{Experiments}
\subsection{Dataset and Validation Scheme}
We use the SEED dataset \cite{zheng2015investigating} to conduct emotion recognition experiments with three affective labels namely negative, neutral, and positive. The stimuli used to collect this dataset were $4$-minute long emotional video clips. Each experiment run consists of $15$ sessions. Each session contains a $5$-second pre-stimuli notice followed by the session-specific visual stimuli. Each session ends with a self-assessment ($45$ seconds) and rest period ($15$ seconds). $15$ subjects ($8$ female and $7$ male) performed the experiments twice, yielding a total of $30$ experiments. EEG were recorded from $62$ electrodes on the scalp (with the international $10-20$ system) at a sampling frequency of $1$ KHz \cite{zheng2015investigating}. In each of the $30$ experiments, we use the first $9$ sessions for training and the remaining $6$ sessions for testing, as pre-defined in \cite{zheng2015investigating}.

\subsection{Implementation Details}
We use the Adam optimizer and its default decay rates for model optimization. We apply gradient clipping with a range of $[-1, 1]$ in order to avoid gradient explosion. The total training epochs are set to $30$ and the learning rate is fixed at $0.001$. All of the experiments are implemented using PyTorch \cite{paszke2019pytorch} on a pair of NVIDIA Tesla P100 GPUs.

\subsection{Expansion to Other AutoEncoders} 
In order to evaluate the generalizability of our method with other forms of autoencoders beyond the proposed recurrent attention-based model, we swap the recurrent autoencoder (plus attention) with an SAE. This variation contains an encoder consisting of 2 FC layers with $256$ and $64$ units with ReLU activation, followed by a decoder with the inverse architecture. The classifier component of the model that learns from the latent representations remains the same. The loss functions and hyperparameters also remain unchanged.

\subsection{Comparison}
We compare our framework with a number of other deep semi-supervised learning techniques \cite{erhan2010does,lee2013pseudo,laine2016temporal,tarvainen2017mean,athiwaratkun2018there,oliver2018realistic,van2020survey}, which were explained in the Related Work section. As these methods have been generally evaluated in domains other than EEG, we re-implement all of them for comparison. The parameters of the models have been designed to achieve maximum performance. In the following sections, we describe the backbone networks used for these benchmarks along with the specifications for each approach.

\textbf{Backbone Networks.}
We employ two deep backbones (DNN and CNN) for the benchmark semi-supervised learning pipelines (except for unsupervised pre-training since it requires an autoencoder). The details of these backbones are as follows. (\textbf{1}) DNN: We use two stacked FC layers ($256$ and $64$ hidden units) followed by the same classifier as the one used in our approach (in Figure \ref{fig: ALSTM-AE}) as the backbone DNN. (\textbf{2}) CNN: We employ two $1$D convolutional layers as the backbone CNN. Specifically, the first convolutional layer has $1$ input channel and $5$ output channels, with kernel size of $3$ and stride of $1$. It is then followed by BatchNorm, LeakyReLU (ratio of $0.3$), and MaxPooling layers. The second convolutional layer has $5$ input channels and $10$ output channels. The kernel size and stride size are kept the same as the first convolutional layer. This convolutional layer is also followed by BatchNorm and LeakyReLU layers to accelerate the training process. Following, a Flatten layer is applied before the classifier.

\textbf{Unsupervised Pre-training.} 
In unsupervised pre-training, we pre-train an SAE as the network as in \cite{erhan2010does} (see the Related Work section). The encoder of the SAE consists of two stacked FC layers ($256$ and $64$ hidden units) and the decoder has the inverse architecture as the encoder. For the DBN approach we employ the pipeline proposed in \cite{xu2016affective}, which consists of $2$ hidden layers ($50$ and $200$ units) followed by a logistic regression layer.

\textbf{Pseudo Labeling.}
Pseudo labeling is performed as in \cite{lee2013pseudo} and explained in the Related Work section. A cross-entropy loss is used for training.

\textbf{$\mathbf{\Pi}$ Model.}
In the $\Pi$ model, we apply a additive Gaussian noise ($\mu=0, \sigma=0.15$) to the input. We do not apply additional dropout since the classifier already contains a dropout rate of $0.5$. Cross-entropy and consistency losses are used for the supervised and unsupervised components respectively. The method is then carried out as in \cite{laine2016temporal} (see the Related Work).

\textbf{Temporal Ensembling.}
Ensemble output ($\mathcal{S}$) is zero initialized in the first epoch. In the remaining epochs, the output aggregation $\mathcal{S}$ is updated by $\mathcal{S} =\delta \mathcal{S} +(1-\delta)s$, where $\delta=0.6$ is the momentum factor and $s$ is the model output in the current epoch $t$. A bias correction operation is further applied on the ensemble output. Consequently, the ensemble output aggregates the model output in previous epochs and updates once per epoch. The remaining settings (e.g., model, input, loss function) are kept the same as in $\Pi$ model \cite{laine2016temporal}.

\textbf{Mean Teacher.}
The student and teacher models share the same architecture and initialized weights. The ensemble model weights are updated using exponential moving average as $\phi_i' = \gamma\phi_{i-1}' + (1-\gamma)\phi_i$, where $i$, $\gamma$, $\phi$, and $\phi'$ denote the training step, smoothing coefficient, and the model weights for the student and teacher \cite{tarvainen2017mean}. We use $\gamma=0.95$ in the experiments. The remaining settings are kept the same as in $\Pi$ model and temporal ensembling \cite{laine2016temporal}.



\section{Results and Analysis}
Table \ref{table} presents the performance of our method in comparison to a number of other SSL benchmarks when all the output labels are used $D_{l} = D$. Here, the results for the pseudo labeling method \cite{lee2013pseudo} are not reported because in this part of our experiments, all the training data have their true labels. It can be observed that the unsupervised pre-training using the SAE performs poorly compared to other methods. When a DNN backbone is used, mean teacher obtains an accuracy of $0.9076\pm0.0772$, which is marginally better than the $\Pi$ model and temporal ensembling. When CNN is used as backbone, the $\Pi$ model achieves an accuracy of $0.8907\pm0.0816$ which slightly outperforms the others. For the $\Pi$ model, temporal ensembling, and mean teacher methods, the performances are consistently better when DNN is used as a backbone. Our framework with the Att. RAE model achieves the highest accuracy of $0.9117\pm 0.0736$.

\begin{table}[!t]
\centering
\setlength\tabcolsep{2 pt}
\caption{The performance of our proposed solution in comparison to other semi-supervised methods when the entire set of output labels are used.} \label{table}

\begin{tabularx}{1.0\columnwidth}{lll}
	\hline
	Paper & Method & Accuracy\\
	\hline\hline
	Erhan et al. \cite{erhan2010does}               & Unsup. Pre-training SAE           & $0.7283 \pm 0.1243 $ \\
    Xu and Plataniotis \cite{xu2016affective}       & Unsup. Pre-training DBN           & $0.7751 \pm 0.0115 $ \\
    Laine and Aila \cite{laine2016temporal}         & $\Pi$ Model  DNN                  & $0.8962 \pm 0.0775 $ \\
    Laine and Aila \cite{laine2016temporal}         & $\Pi$ Model  CNN                  & $0.8907 \pm 0.0816 $ \\
    Laine and Aila \cite{laine2016temporal}         & Temporal Ensembling DNN           & $0.9029 \pm 0.0747 $ \\
	Laine and Aila \cite{laine2016temporal}         & Temporal Ensembling CNN           & $0.8855 \pm 0.0814 $ \\
    Tarvainen and Valpola \cite{tarvainen2017mean}  & Mean Teacher DNN                  & $0.9076 \pm 0.0772 $ \\
    Tarvainen and Valpola \cite{tarvainen2017mean}  & Mean Teacher CNN                  & $0.8756 \pm 0.0872 $ \\
    Ours                                   & SAE                               & $0.9037 \pm 0.0807 $ \\
    Ours                                   & RAE                               & $0.8947 \pm 0.0831 $ \\
    \textbf{Ours}                                   & \textbf{Att. RAE}                          & $0.9117 \pm 0.0736 $ \\
    \hline
\end{tabularx}
\end{table}

\begin{figure}
    \begin{center}
    \includegraphics[width=0.9\columnwidth]{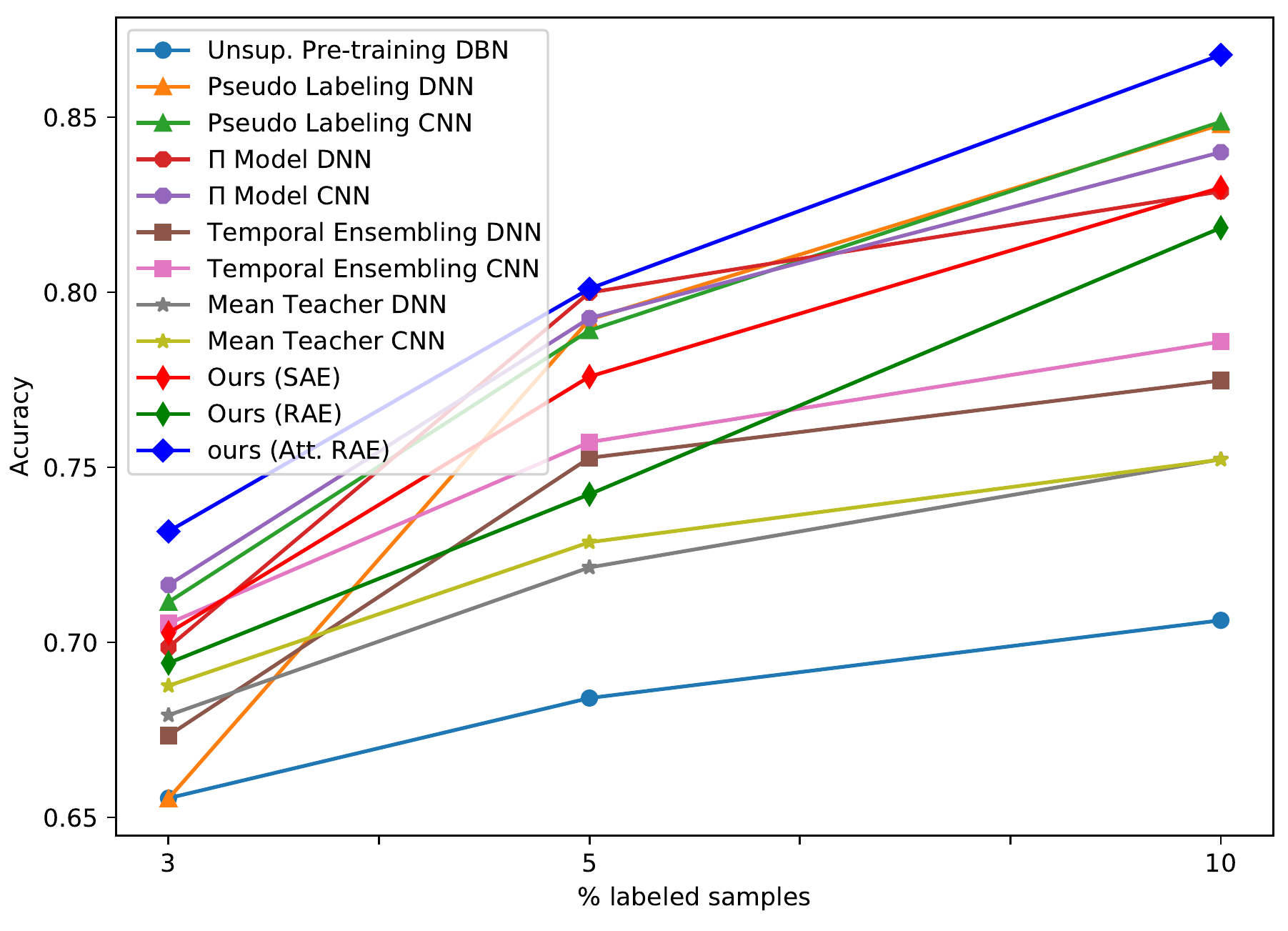} 
    \end{center}
\caption{The average test accuracies obtained by our solution in comparison to other methods when few training samples are labeled ($3\%, 5\%$, and $10\%$).}
\vspace{-2mm}
\label{fig: comparison}
\end{figure}

We also perform experiments where only small subsets ($3\%, 5\%$, and $10\%$) of the ground-truth labels are used as $D_{l} \ll D$, and compare the performance to other semi-supervised learning methods. We ensure that the unlabeled data and labeled data are exactly the same for all the experiments. To show the robustness of our approach, we employ five different random seeds for choice of labeled samples (which are maintained identically throughout all the experiments). Moreover, the evaluation set is always untouched and kept the same for all the experiments. The performances are reported using the averaged results of the experiments repeated with the five different random seeds.

Our experiments indicate that the unsupervised pre-training SAE benchmark obtains relatively low accuracies at $0.4679 \pm 0.1146$, $0.4712 \pm 0.1246 $ and $0.5203 \pm 0.1184$ when $3\%, 5\%$ and $10\%$ samples are labeled. This implies that the representation space obtained by the SAE during unsupervised pre-training lacks sufficient discriminative ability for emotion recognition when labeled data are scarce. Figure \ref{fig: comparison} shows the performance comparison when DNNs and CNNs are used as backbones for the remaining benchmarks (the unsupervised pre-training SAE is not shown to better highlight the differences among the other techniques). As shown in the figure, when only $3\%$ of the samples are labeled and used with a DNN backbone, the $\Pi$ model exhibits an accuracy of $0.6986 \pm 0.1199$, which is better than temporal ensembling ($0.6734\pm0.1104$), mean teacher ($0.6792 \pm 0.0959$), and pseudo labeling ($0.6554 \pm 0.1442$). When $5\%$ of samples are labeled, pseudo labeling and $\Pi$ model achieve results of $0.7922 \pm 0.1312$ and $0.7999 \pm 0.1115$, which are better than temporal ensembling and mean teacher. When more samples are labeled ($10\%$), pseudo labeling obtains an accuracy of $0.8479 \pm 0.1123$, outperforming others.

When only $3\%$ and $5\%$ of the samples are labeled and a CNN is used as the model backbones, the $\Pi$ model achieves accuracies of $0.7164 \pm 0.1011$ and $0.7926 \pm 0.1064$ with better performance than the other benchmarks. When more samples are labeled ($10\%$), pseudo labeling obtains an accuracy of $0.8487 \pm 0.0996$, which is better than others benchmarks. Nonetheless, our framework with Att. RAE consistently achieves the best results ($0.7317 \pm 0.1072$, $0.8010 \pm 0.0938 $, and $0.8678 \pm 0.0890$) when $3\%$, $5\%$, and $10\%$ of training samples are labeled, showing the robustness of our framework. 

Interestingly, when comparing the benchmark models together, we observe that a CNN backbone generally obtains better results for $D_{l} \ll D_{ul}$ (when $3\%$, $5\%$, and $10\%$ of the data are labeled), which is in contrast to our finding when all the data were labeled. This may be due the fact that the non-linear features learned by CNNs from only a few labeled samples could be more representative of the entire feature space.


\begin{figure}[!t]
    \begin{center}
    \includegraphics[width=0.95\columnwidth]{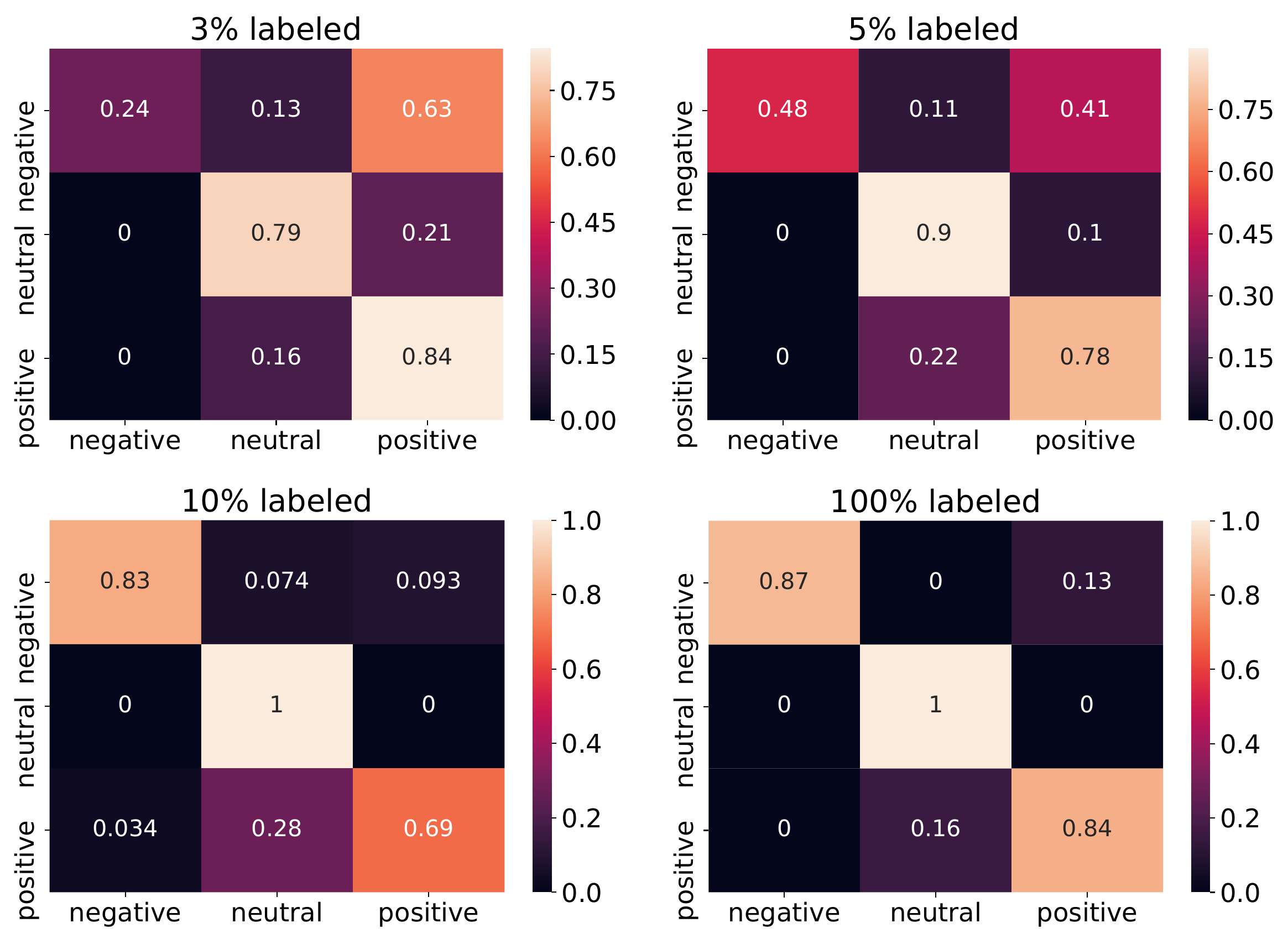}  
    \caption{Confusion matrices for our method (Att. RAE) when different amounts of training samples are labeled (3\%, 5\%, 10\%, and 100\%).}
    \vspace{-2mm}
    \label{fig: confusion_matrix}
    \end{center}
\end{figure}

\begin{figure*}[!t]
    \begin{center}
    \includegraphics[width=2\columnwidth]{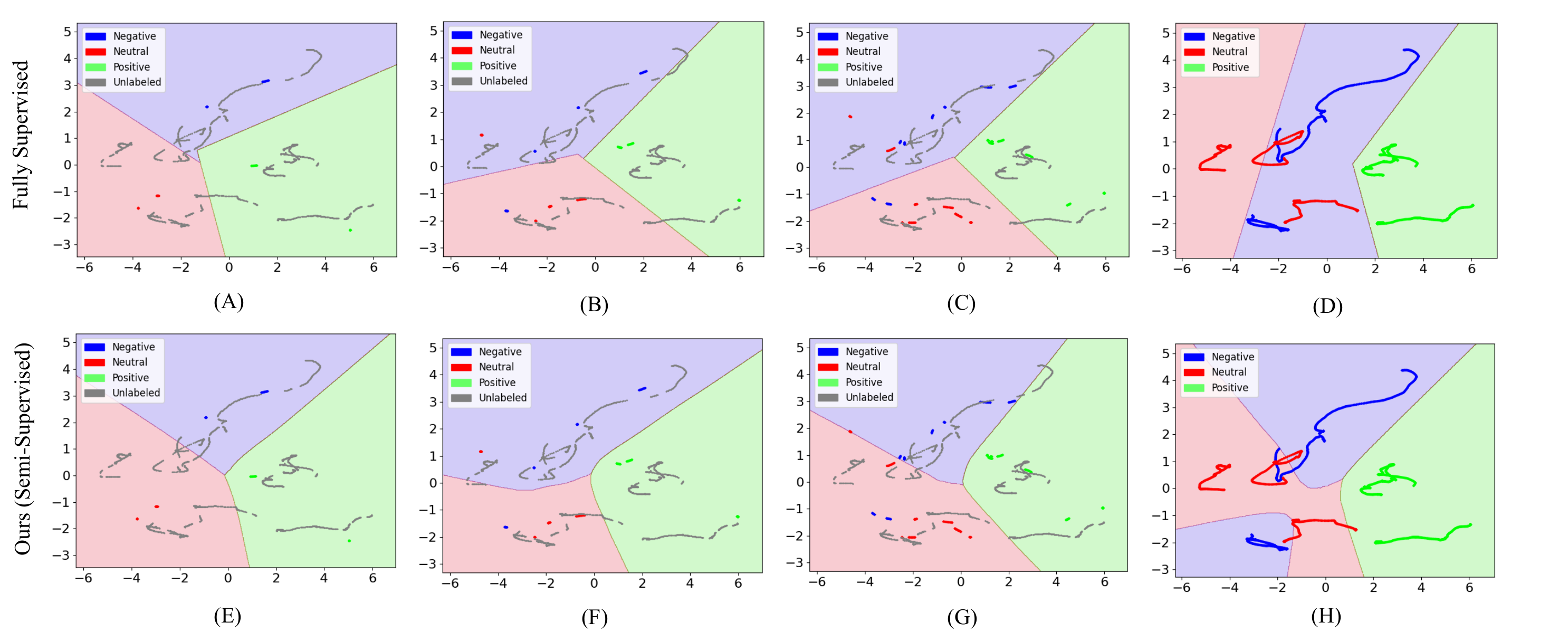}     
    \caption{The decision boundaries of a benchmark fully supervised method (SVM) and our approach (SAE) is depicted when different amounts of training samples are labeled. (A)-(D) are the decision boundaries of the SVM with $3\%$, $5\%$, $10\%$, and $100\%$ labeled training samples,  while (E)-(H) show the decision boundaries of our SAE with $3\%$, $5\%$, $10\%$, and $100\%$ labeled training samples.}
     \vspace{-6mm}
    \label{fig: decision_boundary}
    \end{center}
\end{figure*}


Figure \ref{fig: confusion_matrix} shows the normalized confusion matrix obtained by our approach (Att. RAE), when trained with different amounts of labeled samples. Each row and column present the true labels and predicted labels for each emotion class, respectively. We observe that negative emotion class is generally more difficult to recognize compared to the other two emotion classes, especially when labeled data are extremely low ($3\%$ and $5\%$). Also, the samples of the negative emotion class are frequently misclassified as positive ($0.63$ and $0.41$) when $3\%$ and $5\%$ amounts of training samples are labeled, respectively. In contrast, neutral emotion is much easier to classify compared to the others. Moreover, samples of neutral emotion are misclassified as positive emotion in a few cases, but never misclassified as negative emotion. Both negative and neutral class recognition rates consistently increase when more labeled training samples are available. Especially, the recognition rate of negative emotion increases rapidly. Meanwhile, the recognition rate of positive emotion interestingly stays relatively stable when the size of labeled samples changes.

Here we aim to analyze the difference between our semi-supervised framework vs. a fully supervised solution. To do so, we consider a simple multi-class SVM as a fully supervised method and our proposed framework with an SAE. We then visualize the decision boundaries for when different amounts of labels are using for training in each method. It should be noted that while our semi-supervised solution will leverage the unlabeled portion of the dataset ($D_{ul}$), the fully supervised SVM will not take advantage of $D_{ul}$ and only rely on $D_l$. Figure \ref{fig: decision_boundary} illustrates the decision boundaries obtained using each approach for when $3\%$, $5\%$, $10\%$, and $100\%$ of a training set are labeled. In the figure, the labeled samples for `negative', `neutral', and `positive' are depicted in blue, red, and green respectively, while the unlabeled data are shown in grey. The decision boundaries are specified by the corresponding background colors. In order obtain the decision boundaries in 2D for visualization purposes, we first apply principle component analysis to reduce the EEG features into two a dimensional feature space. Then, we feed the obtained latent features into the two models (our semi-supervised and the fully supervised SVM) to generate the decision boundaries as shown in the figure. Interestingly, we observe that as the amount of labeled samples varies from $3\%$ to $100\%$, the decision boundaries of the fully supervised SVM vary substantially, indicating the naturally the performance is highly dependant on the distribution of labeled data (see Figures \ref{fig: decision_boundary} (A) through (D)). However, the decision boundaries of our semi-supervised framework are much more stable and exhibit considerably less reliance on the amount and distribution of labeled samples. Even when increasing the labels from 10\% to 100\% (Figure \ref{fig: decision_boundary} (G) and (H)), the decision boundaries obtained by our model remain relatively consistent. This observation can be attributed to the fact that our method leverages the unlabeled data when small amounts of labeled samples are used, whereas the fully supervised method does not benefit from such information.

\section{Conclusion}
In this paper, we propose a novel semi-supervised approach for learning EEG representations with reduced dependence on \textit{labeled} samples. Our framework, consisting of a deep attention-based recurrent autoencoder, leverages both large amounts of unlabeled data and few labeled samples for semi-supervised learning as it conducts both unsupervised and supervised learning simultaneously. The unsupervised component maximizes the consistency between original and reconstructed input features for the entire training data (both labeled and unlabeled). Meanwhile the supervised component minimizes the cross-entropy between the input and output labels for the labeled data only. We evaluate our framework on a popular EEG-based emotion recognition dataset and compare our approach with high quality deep semi-supervised benchmarks, demonstrating that it consistently outperforms other methods when few labeled samples are available ($3\%$, $5\%$ and $10\%$). The results show that our approach is capable of learning strong discriminative representations by jointly learning from small amounts of labeled samples and large amounts of unlabeled samples. Further analysis shows that that the decision boundaries obtained by our framework are more stable and less sensitive to limited labeled samples compared to a benchmark fully supervised method.

\bibliographystyle{IEEEtran}
\bibliography{IEEEabrv,ref}

\begin{thebibliography}{10}
\providecommand{\url}[1]{#1}
\csname url@samestyle\endcsname
\providecommand{\newblock}{\relax}
\providecommand{\bibinfo}[2]{#2}
\providecommand{\BIBentrySTDinterwordspacing}{\spaceskip=0pt\relax}
\providecommand{\BIBentryALTinterwordstretchfactor}{4}
\providecommand{\BIBentryALTinterwordspacing}{\spaceskip=\fontdimen2\font plus
\BIBentryALTinterwordstretchfactor\fontdimen3\font minus
  \fontdimen4\font\relax}
\providecommand{\BIBforeignlanguage}[2]{{%
\expandafter\ifx\csname l@#1\endcsname\relax
\typeout{** WARNING: IEEEtran.bst: No hyphenation pattern has been}%
\typeout{** loaded for the language `#1'. Using the pattern for}%
\typeout{** the default language instead.}%
\else
\language=\csname l@#1\endcsname
\fi
#2}}
\providecommand{\BIBdecl}{\relax}
\BIBdecl

\bibitem{dalgleish2004emotional}
T.~Dalgleish, ``The emotional brain,'' \emph{Nature Reviews Neuroscience},
  vol.~5, no.~7, pp. 583--589, 2004.

\bibitem{panksepp2004affective}
J.~Panksepp, \emph{Affective Neuroscience: The Foundations of Human and Animal
  Emotions}.\hskip 1em plus 0.5em minus 0.4em\relax Oxford university press,
  2004.

\bibitem{picard2000affective}
R.~W. Picard, \emph{Affective Computing}.\hskip 1em plus 0.5em minus
  0.4em\relax MIT press, 2000.

\bibitem{jerritta2011physiological}
S.~Jerritta, M.~Murugappan, R.~Nagarajan, and K.~Wan, ``Physiological signals
  based human emotion recognition: a review,'' in \emph{2011 IEEE 7th
  International Colloquium on Signal Processing and its Applications}, 2011,
  pp. 410--415.

\bibitem{sarkar2019classification}
P.~Sarkar, K.~Ross, A.~J. Ruberto, D.~Rodenburg, P.~Hungler, and A.~Etemad,
  ``Classification of cognitive load and expertise for adaptive simulation
  using deep multitask learning,'' \emph{arXiv preprint arXiv:1908.00385},
  2019.

\bibitem{ross2019toward}
K.~Ross, P.~Sarkar, D.~Rodenburg, A.~Ruberto, P.~Hungler, A.~Szulewski,
  D.~Howes, and A.~Etemad, ``Toward dynamically adaptive simulation: Multimodal
  classification of user expertise using wearable devices,'' \emph{Sensors},
  vol.~19, no.~19, p. 4270, 2019.

\bibitem{zhang2019classification}
G.~Zhang, V.~Davoodnia, A.~Sepas-Moghaddam, Y.~Zhang, and A.~Etemad,
  ``Classification of hand movements from eeg using a deep attention-based lstm
  network,'' \emph{IEEE Sensors Journal}, vol.~20, no.~6, pp. 3113--3122, 2019.

\bibitem{zhang2020rfnet}
G.~Zhang and A.~Etemad, ``Rfnet: Riemannian fusion network for eeg-based
  brain-computer interfaces,'' \emph{arXiv preprint arXiv:2008.08633}, 2020.

\bibitem{zhang2019capsule}
G.~Zhang and A.~Etemad, ``Capsule attention for multimodal eeg and eog
  spatiotemporal representation learning with application to driver vigilance
  estimation,'' \emph{arXiv preprint arXiv:1912.07812}, 2019.

\bibitem{zheng2015investigating}
W.-L. Zheng and B.-L. Lu, ``Investigating critical frequency bands and channels
  for eeg-based emotion recognition with deep neural networks,'' \emph{IEEE
  Transactions on Autonomous Mental Development}, vol.~7, no.~3, pp. 162--175,
  2015.

\bibitem{alarcao2017emotions}
S.~M. Alarcao and M.~J. Fonseca, ``Emotions recognition using eeg signals: A
  survey,'' \emph{IEEE Transactions on Affective Computing}, vol.~10, no.~3,
  pp. 374--393, 2017.

\bibitem{zheng2016multichannel}
W.~Zheng, ``Multichannel eeg-based emotion recognition via group sparse
  canonical correlation analysis,'' \emph{IEEE Transactions on Cognitive and
  Developmental Systems}, vol.~9, no.~3, pp. 281--290, 2016.

\bibitem{li2019novel}
Y.~Li, W.~Zheng, L.~Wang, Y.~Zong, L.~Qi, Z.~Cui, T.~Zhang, and T.~Song, ``A
  novel bi-hemispheric discrepancy model for eeg emotion recognition,''
  \emph{arXiv preprint arXiv:1906.01704}, 2019.

\bibitem{tripathi2017using}
S.~Tripathi, S.~Acharya, R.~D. Sharma, S.~Mittal, and S.~Bhattacharya, ``Using
  deep and convolutional neural networks for accurate emotion classification on
  deap dataset,'' in \emph{Twenty-Ninth IAAI Conference}, 2017.

\bibitem{cho2014learning}
K.~Cho, B.~Van~Merri{\"e}nboer, C.~Gulcehre, D.~Bahdanau, F.~Bougares,
  H.~Schwenk, and Y.~Bengio, ``Learning phrase representations using {RNN}
  encoder-decoder for statistical machine translation,'' \emph{arXiv preprint
  arXiv:1406.1078}, 2014.

\bibitem{sepas2020long}
A.~Sepas-Moghaddam, A.~Etemad, F.~Pereira, and P.~L. Correia, ``Long short-term
  memory with gate and state level fusion for light field-based face
  recognition,'' \emph{IEEE Transactions on Information Forensics and
  Security}, vol.~16, pp. 1365--1379, 2020.

\bibitem{zhang2016continuous}
N.~Zhang, W.-L. Zheng, W.~Liu, and B.-L. Lu, ``Continuous vigilance estimation
  using lstm neural networks,'' in \emph{International Conference on Neural
  Information Processing}.\hskip 1em plus 0.5em minus 0.4em\relax Springer,
  2016, pp. 530--537.

\bibitem{zhang2018spatial}
T.~Zhang, W.~Zheng, Z.~Cui, Y.~Zong, and Y.~Li, ``Spatial-temporal recurrent
  neural network for emotion recognition,'' \emph{IEEE Transactions on
  Cybernetics}, no.~99, pp. 1--9, 2018.

\bibitem{song2018eeg}
T.~Song, W.~Zheng, P.~Song, and Z.~Cui, ``Eeg emotion recognition using
  dynamical graph convolutional neural networks,'' \emph{IEEE Transactions on
  Affective Computing}, 2018.

\bibitem{zhong2020eeg}
P.~Zhong, D.~Wang, and C.~Miao, ``Eeg-based emotion recognition using
  regularized graph neural networks,'' \emph{IEEE Transactions on Affective
  Computing}, 2020.

\bibitem{bulat2021pre}
A.~Bulat, S.~Cheng, J.~Yang, A.~Garbett, E.~Sanchez, and G.~Tzimiropoulos,
  ``Pre-training strategies and datasets for facial representation learning,''
  \emph{arXiv preprint arXiv:2103.16554}, 2021.

\bibitem{erhan2010does}
D.~Erhan, A.~Courville, Y.~Bengio, and P.~Vincent, ``Why does unsupervised
  pre-training help deep learning?'' in \emph{Proceedings of the thirteenth
  international conference on artificial intelligence and statistics}.\hskip
  1em plus 0.5em minus 0.4em\relax JMLR Workshop and Conference Proceedings,
  2010, pp. 201--208.

\bibitem{van2020survey}
J.~E. Van~Engelen and H.~H. Hoos, ``A survey on semi-supervised learning,''
  \emph{Machine Learning}, vol. 109, no.~2, pp. 373--440, 2020.

\bibitem{xu2016affective}
H.~Xu and K.~N. Plataniotis, ``Affective states classification using eeg and
  semi-supervised deep learning approaches,'' in \emph{2016 IEEE 18th
  International Workshop on Multimedia Signal Processing (MMSP)}.\hskip 1em
  plus 0.5em minus 0.4em\relax IEEE, 2016, pp. 1--6.

\bibitem{lee2013pseudo}
D.-H. Lee \emph{et~al.}, ``Pseudo-label: The simple and efficient
  semi-supervised learning method for deep neural networks,'' in \emph{Workshop
  on challenges in representation learning, ICML}, vol.~3, no.~2, 2013.

\bibitem{oliver2018realistic}
A.~Oliver, A.~Odena, C.~Raffel, E.~D. Cubuk, and I.~J. Goodfellow, ``Realistic
  evaluation of deep semi-supervised learning algorithms,'' \emph{arXiv
  preprint arXiv:1804.09170}, 2018.

\bibitem{laine2016temporal}
S.~Laine and T.~Aila, ``Temporal ensembling for semi-supervised learning,''
  \emph{arXiv preprint arXiv:1610.02242}, 2016.

\bibitem{athiwaratkun2018there}
B.~Athiwaratkun, M.~Finzi, P.~Izmailov, and A.~G. Wilson, ``There are many
  consistent explanations of unlabeled data: Why you should average,''
  \emph{arXiv preprint arXiv:1806.05594}, 2018.

\bibitem{tarvainen2017mean}
A.~Tarvainen and H.~Valpola, ``Mean teachers are better role models:
  Weight-averaged consistency targets improve semi-supervised deep learning
  results,'' \emph{arXiv preprint arXiv:1703.01780}, 2017.

\bibitem{li2020extraction}
B.~Li and A.~Sano, ``Extraction and interpretation of deep autoencoder-based
  temporal features from wearables for forecasting personalized mood, health,
  and stress,'' \emph{Proceedings of the ACM on Interactive, Mobile, Wearable
  and Ubiquitous Technologies}, vol.~4, no.~2, pp. 1--26, 2020.

\bibitem{paszke2019pytorch}
A.~Paszke, S.~Gross, F.~Massa, A.~Lerer, J.~Bradbury, G.~Chanan, T.~Killeen,
  Z.~Lin, N.~Gimelshein, L.~Antiga \emph{et~al.}, ``Pytorch: An imperative
  style, high-performance deep learning library,'' \emph{arXiv preprint
  arXiv:1912.01703}, 2019.

\end{thebibliography}

\end{document}